%% file: main.tex
\pdfoutput=1
\documentclass[11pt]{article}
\usepackage{naacl2021}

\input{dlab_macros}
\usepackage{times}
\usepackage{latexsym}
\usepackage{booktabs}
\usepackage{multirow}
\usepackage{graphicx}
\usepackage{subfig}
\usepackage{amsmath}

\usepackage{booktabs} 
\usepackage{threeparttable}
\usepackage{enumitem}
\usepackage{xspace}
\usepackage[bottom]{footmisc}

\newcommand{\ignore}[1]{}

\newcommand{\tabcaption}[1]{\vspace*{-3mm}\caption{#1}\vspace*{-4mm}}

\newcommand{\moveup}{\vspace*{-2mm}}
\newcommand{\moveups}{\vspace*{-1mm}}

\newcommand{\qb}{\textsc{Quotebank}\xspace}
\newcommand{\ethemes}{\textsc{Eigenthemes}\xspace}
\newcommand{\genre}{\textsc{GENRE}\xspace}
\definecolor{teal}{rgb}{0.07, 0.04, 0.56} 

\usepackage[T1]{fontenc}
\usepackage[utf8]{inputenc}
\usepackage{microtype}

\begin{document}

\title{Strong Heuristics for Named Entity Linking}

\newcommand{\zagreb}{$^\ddagger$}
\newcommand{\konstanz}{$^\mathsection$}
\newcommand{\epfl}{$^\mathparagraph$}

\author{
  Marko \v{C}uljak,\zagreb{}\thanks{~~Research done while at EPFL.}~ Andreas Spitz,\konstanz{} Robert West,\epfl{} Akhil Arora\epfl{}\thanks{~~Corresponding author.} \\
  \zagreb University of Zagreb, \konstanz University of Konstanz, \epfl EPFL \\
\href{mailto:marko.culjak@fer.hr}{\texttt{{ marko.culjak@fer.hr}}}, \href{mailto:andreas.spitz@uni-konstanz.de}{\texttt{{ andreas.spitz@uni-konstanz.de}}} \\
\href{mailto:robert.west@epfl.ch}{\texttt{{ robert.west@epfl.ch}}}, 
\href{mailto:akhil.arora@epfl.ch}{\texttt{{ akhil.arora@epfl.ch}}}
}

\maketitle


\begin{abstract}

Named entity linking (NEL) in news is a challenging endeavour due to the frequency of unseen and emerging entities, which necessitates the use of unsupervised or zero-shot methods. However, such methods tend to come with caveats, such as no integration of suitable knowledge bases (like Wikidata) for emerging entities, a lack of scalability, and poor interpretability. Here, we consider person disambiguation in \qb, a massive corpus of speaker-attributed quotations from the news, and investigate the suitability of intuitive, lightweight, and scalable heuristics for NEL in web-scale corpora. Our best performing heuristic disambiguates 94\% and 63\% of the mentions on \qb and the AIDA-CoNLL benchmark, respectively. Additionally, the proposed heuristics compare favourably to the state-of-the-art unsupervised and zero-shot methods, \ethemes and m\genre, respectively, thereby serving as strong baselines for unsupervised and zero-shot entity linking.

\end{abstract}


\section{Introduction}

While many of the most famous historic quotes are wise irrespective of their origin, this is less true for the majority of contemporary quotes in the news, which require speaker attribution to be useful in journalism or the social and political sciences. This observation is the motivation behind the construction of \qb, a corpus of 178 million unique quotations that are attributed to speaker mentions and were extracted from 162 million news articles published between 2008 and 2020~\cite{quotebank}. However, given the ambiguity of names, attributing quotes to mentions is insufficient for proper attribution, and thus, named entity disambiguation is required, a feature which \qb lacks.

To tackle this shortcoming and investigate the disambiguation of person mentions in \qb as a prototypical example of a web-scale corpus, we explore the suitability of scalable named entity linking (NEL) heuristics, which map mentions of entity names in the text to a unique identifier in a referent knowledge base (KB) and thus, resolve the ambiguity. NEL is an established task and solutions have been used for a variety of applications such as KB population \cite{dredze-2010-entity} or information extraction \cite{hoffart-etal-2011-robust}, yet the frequency of emerging and unseen entities in news data renders the adaptation of supervised NEL approaches difficult and tends to require unsupervised or zero-shot methods. 

While such unsupervised methods \cite{le2019distant, arora-etal-2021-low} and zero-short methods  \cite{logeswaran-etal-2019-zero, DeCao2021AutoregressiveER} have been developed in recent years, scalability is an issue. For example, fully disambiguating \qb with the state-of-the-art zero-shot NEL method, m\genre \cite{DeCao2021MultilingualAE}, would require approximately 37 years on a single GPU according to our experimental estimates. Therefore, we investigate the suitability of heuristic NEL methods that rely on signals that are simple to extract from mention contexts or entity entries in a KB. In contrast to m\genre, we find that our best-performing heuristics can solve the same task in 108 days on a single CPU core, \ie, orders of magnitude faster and on cheaper hardware, while achieving comparable performance.

\xhdr{Contributions}
To address the need for NEL in web-scale corpora, we investigate the disambiguation performance of simple, interpretable, scalable, and lightweight heuristics and compare them to state-of-the-art zero-shot and unsupervised NEL methods. Our experiments on \qb and the AIDA-CoNLL benchmark demonstrate the competitiveness of these heuristics.


\section{Related Work}

Viable learning-based methods for NEL in settings without available training data can be classified into \emph{zero-shot} and \emph{unsupervised} learning.

\xhdrNoPeriod{Zero-shot NEL} was introduced by \citet{logeswaran-etal-2019-zero} with the objective of linking mentions to entities that were unseen during training. Later, \citet{wu2019scalable} proposed a BERT-based model for this task. Finally, \citet{DeCao2021AutoregressiveER} proposed \genre, a supervised NEL method that leverages BART to retrieve entities by generating their unique names autoregressively, conditioned on the context by employing beam search. While \genre uses Wikipedia as its referent KB and is not directly compatible with our setting, we compare our methods to m\genre \cite{DeCao2021MultilingualAE}, a multilingual adaptation of \genre using Wikidata.

\xhdr{Unsupervised NEL} 
\citet{le2019distant} proposed $\tau$MIL-ND, a BiLSTM model trained on noisy labels, which are generated via a heuristic that ranks the candidate entities of a mention based on matching words in a mention and candidate labels. 
Similarly, \citet{fan2015distant} experiment with distant learning for NEL and create training data by merging Freebase with Wikipedia. 
Recently, \citet{arora-etal-2021-low} proposed \ethemes, which is based on the observation that vector representations of gold entities lie in a low-rank subspace of the full embedding space. These low-rank subspaces are used to perform collective entity disambiguation.

While powerful, the aforementioned methods are designed for general domains and multiple entity types, and thus, cannot capitalize on domain- and entity-specific signals. In the following, we investigate the suitability of unsupervised NEL heuristics for person disambiguation in the domain of news quotes in comparison to these methods.


\section{Problem Formalization}
\label{sec:problem}

The input to our NEL system are articles $a\in \mathcal{A}$ from the set $\mathcal A$ of all articles in \qb. In each article $a$, a set of entity mentions $\mathcal M_a$ is annotated. Each such mention $m \in \mathcal{M}_a$ can be mapped to a set of candidate Wikidata entities $\mathcal E_m$, which are uniquely identified by their Wikidata QID identifier (for further details regarding Wikidata, see Appendix~\ref{app:wiki}). 
If multiple entity candidates are available for a mention, we refer to this mention as \emph{ambiguous}. Conversely, \emph{unambiguous} mentions have only a single candidate entity. Given an article $a\in \mathcal{A}$, an ambiguous mention $m \in \mathcal M_a$, and all candidate entities $\mathcal E_m$, the task of NEL is to identify the entity $e \in \mathcal E_m$ to which $m$ refers. 

We assume that NEL methods assign a rank $r(e,m)$ to each candidate entity $e \in \mathcal{E}_m$ by ranking candidates according to the score provided by the method, which corresponds to the likelihood that $e$ is the correct entity for $m$. Consequently, we assume that methods cannot identify cases in which the entity does not exist in the KB or is not contained in the list of candidates (i.e., out-of-KB or NIL predictions). Thus, our focus is on the evaluation of methods in cases where at least one candidate is available.


\section{Scoring Methods}
\label{sec:scoring}

We consider three main signals for entity candidate ranking methods: \textit{entity popularity}, \textit{entity-content similarity}, and \textit{entity-entity similarity}. Implementation details are provided in Appendix~\ref{app:methods}. 

\subsection{Entity Popularity}

Entity popularity is an important signal for disambiguating entities in news articles as popular entities are more likely to appear in the news \citep{shen2015entity}. Since popularity cannot be measured directly, we utilize 4 proxies derived from Wikidata, some of which have also been used previously as features for supervised NEL \citep{delpeuch2020opentapioca}.

\xhdr{Number of properties (NP)}
Based on the assumption that Wikidata contains more information for popular entities, we use the number of Wikidata properties to approximate entity popularity.

\xhdr{Number of site links (NS)} 
Similar to NP, a more popular entity is likely connected to more Wikimedia pages. We thus use the number of site links to estimate entity popularity.

\xhdrNoPeriod{PageRank (PR)} 
is a graph centrality metric that was originally developed for web search as a part of Google's search engine \cite{Page1999ThePC}. We experiment with two PageRank scores computed on the Wikidata graph (PR$_{\mathrm{WD}}$) and the Wikipedia graph (PR$_{\mathrm{WP}}$) and report their results separately.

\xhdr{Lowest QID (LQID)} 
The Wikidata QID is an auto-incremented integer identifier. Intuitively, well-known entities are added to Wikidata early and their QIDs are low. Therefore, we simply select the candidate with the lowest QID value.

\subsection{Entity-Content Similarity} 
\label{ssec:ecsim}
In addition to entity-centric information, we consider the mention context and attempt to match it to the attributes of candidate entities in the KB. Consider the following example from \qb:\\[3pt]
\textit{``Professor \textbf{\underline{Tim Wheeler}, Vice-Chancellor of the University of Chester}, said: "The university is dedicated to educating the very best nurses [...]''}\\[3pt] 
Tim Wheeler's title, \textit{Vice-Chancellor of the University of Chester}, exactly matches the short description of a Wikidata entity with QID Q2434362. Therefore, it stands to reason that we can leverage content similarity metrics for entity linking.

\xhdr{Intersection score (IScore)} The IScore captures word overlap between mention context and entity descriptions. Let $\mathcal W_a$ be a set of lowercased words occurring in article $a$, let $\mathcal W_e$ be a set of words occurring in the textual representation of an entity in Wikidata, and let $\mathcal W_{sw}$ be a set of English stopwords. 
We then compute the IScore of an entity $e$ with respect to article $a$ as
 \begin{equation}
    \label{eq:iscore}
    \moveups
     \text{IScore}(a, e) = |(\mathcal W_a \cap \mathcal W_e) \setminus \mathcal W_{sw}|
 \end{equation}
While we could normalize the score by $|\mathcal W_a \cup \mathcal W_e|$ to obtain a Jaccard similarity, we intentionally bias the IScore towards entities with more substantial descriptions, thereby implicitly incorporating entity popularity information. We use the Porter stemmer \citep{porter1980algorithm} for stemming words before matching (please see Appendix~\ref{app:iscoreexp} for experiments with IScore using raw input words or lemmatization).

\xhdr{Narrow IScore (NIScore)}
For a more focused context representation, we also compute a version of the IScore with a narrow context that only contains the sentences in which a mention of the given entity occurs. For further experiments with the selection of mention contexts, see Appendix~\ref{app:iscoreexp}.
 
\xhdr{Cosine similarity of embeddings (CSE)}  Following a baseline from \citet{arora-etal-2021-low}, to capitalize on the effectiveness of transformer models for NLP tasks, we leverage contextualized language models to create embeddings of article contents and candidate entity descriptions, which are then compared. We employ BART$_\text{BASE}$ \citep{lewis-etal-2020-bart} to generate embeddings and then compute cosine similarity scores. For details, see Appendix~\ref{app:methods}.

\xhdr{Narrow CSE (NCSE)}
Similar to the NIScore, we consider a narrow context around entity mentions for computing the CSE by restricting the context that is used for the creation of embeddings to sentences in which the entity occurs.
 
\subsection{Entity-Entity Similarity}
Since many mentions of entities can be expected to be unambiguous, we may use such mentions as anchors and leverage their relations to ambiguous mentions for the purpose of disambiguation. 
Similar to the entity-content similarity methods described above, we experiment with metrics that use intersections of entity occurrences and embedding similarities of attribute values from Wikidata.

\xhdr{Entity-entity IScore (EEIScore)} 
Following the above intuition, the EEIScore utilizes the information that is contained in relations between ambiguous and unambiguous mentions. Let $\mathcal{U}_a$ be the set of all entities that can be mapped to unambiguous mentions in an article $a$ (i.e., mentions that can be trivially disambiguated). Let $\mathcal{S}_e$ be the set of all statements that occur in the Wikidata entry corresponding to an entity $e$. We define $\mathcal{S}_{\mathcal{U}_a} := \bigcup_{e \in \mathcal{U}_a}\mathcal{S}_e$. Using this set of all statements of unambiguous entities, we then compute the EEIScore of a candidate entity $e$ for an ambiguous mention as:
\begin{equation}
    \textrm{EEIScore}(e, \mathcal{U}_a) =  |\mathcal{S}_e \cap \mathcal{S}_{\mathcal{U}_a}|
    \moveup
\end{equation}

\xhdr{Cosine similarity of statement value embeddings (CSSVE)} 
We refine the idea behind the intersection score of entity relations by using embeddings of Wikidata statement values and property types (i.e., relations in Wikidata). For each entity $e$, Wikidata contains a set of statements $s_e = (p_e, v_e)$, consisting of a property $p_e$ and a value $v_e$. Using this data, we first create embeddings $\varepsilon(v)$ of the values for all statements $s\in \mathcal{S}_{\mathcal{U}_a}\cup \mathcal S_e$. We then compute CSSVE as the sum of cosine similarities of statement value embeddings between all pairs of statements of the candidate entity and statements of unambiguous mentions in the article that have matching property types (i.e., describe the same type of relation):
\begin{equation}
\small
\textrm{CSSVE}(e, \mathcal{U}_a) = \sum_{\substack{(s_u, s_e) \in (\mathcal S_{\mathcal{U}_a} \times \mathcal S_e)\\ p_u = p_e}} \frac{ \varepsilon(v_u) \cdot \varepsilon(v_e) }{\| \varepsilon(v_u) \|\| \varepsilon(v_e) \|}
\end{equation}

\subsection{Composite Scores}
We also use two composite scores in our evaluation: \textbf{UIScore} refers to the weighted sum of IScore, NIScore, and EEIScore, while \textbf{UCSE} refers to the weighted sum of CSE, NCSE, and CSSVE. Since CSE and NCSE are cosine similarities, their outputs are constrained to the $[-1,1]$ interval, while CSSVE is unbounded. To ensure similar magnitudes we map all scores to the $[0, 1]$ interval by applying the transformation $f(x) = \frac{1}{2}(x + 1)$ to CSE and NCSE, and additive smoothing to CSSVE.


\section{Data}
\label{ambiguity}

We focus on \qb data, but also investigate the performance on AIDA-CoNLL as a benchmark. Similar to \citealt{arora-etal-2021-low}, \citealt{deeptype}, and \citealt{Guo2018RobustNE} we label the mentions as either \emph{`easy'} or \emph{`hard'}. In \qb, we deem a mention easy if it can be correctly disambiguated using NS and hard otherwise, while in AIDA-CoNLL we use the definition proposed by \citealt{arora-etal-2021-low}. In Table \ref{tab:data-stats} we present the statistics for easy and hard mentions in the datasets.

\xhdrNoPeriod{\qb} is a collection of quotes that were extracted from $127$ million news articles and attributed to one of $575$ million speaker mentions \cite{quotebank}, out of which 75\% are unambiguous. For our evaluation, we use a randomly sampled subset of $300$ articles that are manually annotated with $1{,}866$ disambiguated person mentions. 70\% of these mentions are unambiguous. Out of the ambiguous mentions, it was possible to determine ground truth labels for 310 (57\%), which we use in our evaluation. We split the ground truth into 245 mentions (79\%) for evaluation and 65 mentions (21\%) for parameter tuning. For a more thorough description of the \qb ground truth, see Appendix~\ref{app:gt}.

\xhdr{AIDA-CoNLL} 
To assess whether the proposed methods can be used for unsupervised NEL in general, we also evaluate their performance on the AIDA-CoNLL benchmark \citep{hoffart-etal-2011-robust}, which is based on the CoNLL 2003 shared task \cite{DBLP:conf/conll/SangM03}. We use the same setup as \citealt{arora-etal-2021-low} and use the validation set for hyperparameter optimization. The differences between the evaluation setups of \qb and AIDA-CoNLL are explained in Appendix \ref{app:qb-aida-diffs}.

\begin{table}[ht]
\caption{The number of mentions in different difficulty categories. The definitions of \textit{Easy} and \textit{Hard} mentions are presented in \S~\ref{sec:results}. On AIDA-CoNLL, \#Easy + \#Hard $\neq$ \#Overall because for some mentions, the gold-entity was not contained in the candidate set.}
\begin{tabular}{lccc}
\hline
Dataset    & \#Easy & \#Hard & \#Overall \\ \midrule
\qb  & 203    & 42     & 245       \\
AIDA-CoNLL & 2555   & 1136   & 4478      \\ \midrule
\end{tabular}
\moveup
\moveup
\moveups
\label{tab:data-stats}
\end{table}


\section{Evaluation}
\label{sec:res}

All the resources (code, datasets, \etc) required to reproduce the experiments in this paper are available at~\url{https://github.com/epfl-dlab/nelight}.

\subsection{Evaluation Setup}

We use \textit{micro} precision at one (P@1) and mean reciprocal rank (MRR) as the evaluation metrics. The metrics are aggregated over all ambiguous mentions for which ground truth data is available. Performance is reported with 95\% bootstrapped confidence intervals (CIs) over $10,000$ bootstrap samples. To identify optimal weight parameters for the composite metrics, we perform a grid search over the range $[0, 1]$. For the \qb data, the best performance is obtained for weights $(1, 1, 1)$ for UIScore and $(0.45, 0.9, 0.2)$ for UCSE. For the AIDA-CoNLL data, we perform the parameter optimization on the official validation set, where the best performance is obtained for weights $(0.9, 0, 1)$ and $(0, 1, 1)$ for UIScore and UCSE, respectively.

\xhdr{Tie breaking} Several ranking methods introduce ties, which we break by using popularity heuristics. Among the popularity heuristics, only LQID is injective and always outputs distinct scores for different entities. In our experiments, we, therefore, use LQID to break ties if they remain after using other tie-breakers. A full breakdown of the tie-breaking performance for all popularity-based methods can be found in Appendix~\ref{app:tie}.

\subsection{Results}
\label{sec:results}
We report P@1 for all the methods in Table~\ref{tab:main-res}, and MRR in Appendix \ref{app:main-mrr}. For comparison, we present the analytically computed performance of a random baseline, which picks one of the entity candidates uniformly at random.

\begin{table*}[t]
\tabcaption{P@1 of the methods on \qb and AIDA-CoNLL.  Eigen (IScore) refers to \ethemes weighted by IScore. Eigen on \qb is weighted by NS, while on AIDA, it denotes the results obtained by \citealt{arora-etal-2021-low}.  The best obtained P@1 in each column is highlighted \textbf{bold}.}
\label{tab:main-res}
\vspace*{2mm}
\centering
\begin{threeparttable}
\resizebox{\textwidth}{!}{
\begin{tabular}{lcccccc}
\hline
                   & \multicolumn{3}{c}{\qb}                                                        & \multicolumn{3}{c}{AIDA-CoNLL}                                                       \\ \cmidrule(lr){2-4}\cmidrule(lr){5-7} 
\textbf{Method}    & \textbf{Easy}              & \textbf{Hard}              & \textbf{Overall}           & \textbf{Easy}              & \textbf{Hard}              & \textbf{Overall}           \\ \hline
Random             & 0.374 $\pm$ 0.017          & 0.260 $\pm$ 0.045          & 0.354 $\pm$ 0.024          & 0.267 $\pm$ 0.014       & 0.066 $\pm$ 0.004          & 0.169 $\pm$ 0.009          \\ \hline
LQID               & 0.828 $\pm$ 0.054          & 0.238 $\pm$ 0.140          & 0.727 $\pm$ 0.056          & 0.856 $\pm$ 0.014          & 0.259 $\pm$ 0.029          & 0.554 $\pm$ 0.016          \\
NP                 & 0.921 $\pm$ 0.040          & 0.143 $\pm$ 0.120          & 0.788 $\pm$ 0.052          & 0.856 $\pm$ 0.014          & 0.190 $\pm$ 0.023          & 0.536 $\pm$ 0.015          \\
NS                 & \textbf{1.000 $\pm$ 0.000} & 0.000 $\pm$ 0.000          & 0.829 $\pm$ 0.048          & 0.908 $\pm$ 0.012          & 0.275 $\pm$ 0.026          & 0.588 $\pm$ 0.014          \\
PR$_{\mathrm{WD}}$ & 0.768 $\pm$ 0.059          & 0.214 $\pm$ 0.132          & 0.673 $\pm$ 0.061          & 0.838 $\pm$ 0.014          & 0.155 $\pm$ 0.021          & 0.517 $\pm$ 0.015          \\
PR$_{\mathrm{WP}}$ & 0.926 $\pm$ 0.040          & 0.333 $\pm$ 0.140          & 0.824 $\pm$ 0.048          & \textbf{0.938 $\pm$ 0.010} & 0.282 $\pm$ 0.027          & 0.607 $\pm$ 0.014          \\ \hline
IScore             & 0.956 $\pm$ 0.030          & 0.762 $\pm$ 0.134          & 0.922 $\pm$ 0.034          & 0.863 $\pm$ 0.014          & 0.549 $\pm$ 0.029          & 0.632 $\pm$ 0.015          \\
NIScore            & 0.966 $\pm$ 0.030          & 0.571 $\pm$ 0.151          & 0.851 $\pm$ 0.014          & 0.851 $\pm$ 0.014          & 0.407 $\pm$ 0.028          & 0.562 $\pm$ 0.015          \\
CSE                & 0.901 $\pm$ 0.044          & 0.500 $\pm$ 0.159          & 0.833 $\pm$ 0.047          & 0.386 $\pm$ 0.019          & 0.276 $\pm$ 0.026          & 0.290 $\pm$ 0.014          \\ \hline
EEIScore           & 0.951 $\pm$ 0.034          & 0.690 $\pm$ 0.143          & 0.906 $\pm$ 0.036          & 0.815 $\pm$ 0.016          & 0.382 $\pm$ 0.031          & 0.562 $\pm$ 0.015          \\
CSSVE              & 0.872 $\pm$ 0.049          & 0.357 $\pm$ 0.155          & 0.784 $\pm$ 0.051          & 0.712 $\pm$ 0.017          & 0.256 $\pm$ 0.026          & 0.471 $\pm$ 0.015          \\ \hline
UIScore            & 0.966 $\pm$ 0.030          & \textbf{0.833 $\pm$ 0.123} & 0.943 $\pm$ 0.029          & 0.833 $\pm$ 0.014          & 0.577 $\pm$ 0.028          & 0.621 $\pm$ 0.014          \\
UCSE               & 0.941 $\pm$ 0.034          & 0.595 $\pm$ 0.156          & 0.882 $\pm$ 0.042          & 0.465 $\pm$ 0.019          & 0.386 $\pm$ 0.029          & 0.363 $\pm$ 0.014          \\ \hline
Eigen              & 0.995 $\pm$ 0.010          & 0.238 $\pm$ 0.134          & 0.865 $\pm$ 0.044          & 0.859 $\pm$ 0.014          & 0.500 $\pm$ 0.030          & 0.617 $\pm$ 0.015          \\
Eigen (IScore)      & 0.956 $\pm$ 0.030          & 0.714 $\pm$ 0.147          & 0.914 $\pm$ 0.037          & 0.794 $\pm$ 0.015          & \textbf{0.702 $\pm$ 0.029}\tnote{\dag} & 0.631 $\pm$ 0.014          \\
m\genre             & 0.995 $\pm$ 0.010          & 0.810 $\pm$ 0.143          & \textbf{0.963 $\pm$ 0.025} & 0.925 $\pm$ 0.011          & 0.610 $\pm$ 0.028          & \textbf{0.682 $\pm$ 0.014}\tnote{\dag} \\
\hline
\end{tabular}}
\begin{tablenotes}
\small \item[\dag] Indicates statistical significance ($p<0.05$) between the best and the second-best method using bootstrapped 95\% CIs.
\end{tablenotes}
\end{threeparttable}
\moveup
\end{table*}

\xhdr{\qb}
Among the popularity-based metrics, the best results are achieved by NS. However, considering the confidence intervals, the performance gains of NS over PR$_{\mathrm{WP}}$ and NP are not significant. LQID and PR$_{\mathrm{WD}}$ perform poorly in comparison to the other methods. All popularity methods outperform the random baseline, confirming their usefulness as a prior for NEL.

\begin{table}[t]
\centering
\caption{P@1 of representative methods on various entity types in the AIDA-CoNLL dataset. In the evaluation dataset, there are 1016 PER, 1345 ORG, 1575 LOC, and 542 MISC mentions. The best P@1 in each column is highlighted \textbf{bold}.}
\label{tab:entity-type}
\resizebox{.99\linewidth}{!}{\begin{tabular}{lccccc}
\hline
\textbf{Method}    & \textbf{PER}               & \textbf{ORG}               & \textbf{LOC}               & \textbf{MISC}              \\ \hline
NS                 & 0.687 $\pm$ 0.030          & 0.410 $\pm$ 0.027          & 0.777 $\pm$ 0.021          & 0.292 $\pm$ 0.039          \\
PR$_{\mathrm{WP}}$ & 0.719 $\pm$ 0.029          & 0.477 $\pm$ 0.026          & 0.752 $\pm$ 0.023          & \textbf{0.293 $\pm$ 0.042} \\ \hline
IScore             & 0.786 $\pm$ 0.026          & 0.597 $\pm$ 0.026          & 0.694 $\pm$ 0.022          & 0.245 $\pm$ 0.035          \\
UIScore            & \textbf{0.789 $\pm$ 0.026} & 0.601 $\pm$ 0.026          & 0.664 $\pm$ 0.023          & 0.232 $\pm$ 0.035          \\ \hline
m\genre             & 0.720 $\pm$ 0.027          & 0.608 $\pm$ 0.027          & \textbf{0.858 $\pm$ 0.018} & 0.284 $\pm$ 0.039          \\
Eigen (IScore)     & 0.760 $\pm$ 0.026          & \textbf{0.732 $\pm$ 0.025} & 0.608 $\pm$ 0.024          & 0.205 $\pm$ 0.035          \\
Eigen              & 0.696 $\pm$ 0.028          & 0.671 $\pm$ 0.026          & 0.655 $\pm$ 0.023          & 0.223 $\pm$ 0.035         \\ \bottomrule
\end{tabular}} 
\moveup
\moveup
\moveups
\end{table}

The performances of entity-entity similarity methods are similar to their entity-content similarity counterparts. This is in line with the hypothesis that the gold entities mentioned in the same article are more closely related than the other subsets of entity candidates \cite{arora-etal-2021-low}. Generally, combining the entity-content similarity methods with their entity-entity similarity counterparts leads to performance gain, as seen from the example of UIScore and UCSE. Considering the overall performance, UIScore outperforms CSE and all entity popularity methods. The performance of CSE is similar to the performance of NS, which is considerably simpler.  Finally, the performance of UIScore is comparable to m\genre, achieving a slightly higher P@1 on hard mentions.

\xhdr{AIDA-CoNLL} In the AIDA-CoNLL data, IScore and UIScore achieve a comparable performance to the current state-of-the-art in unsupervised entity linking, \ethemes \cite{arora-etal-2021-low}, but lag slightly behind m\genre \cite{DeCao2021MultilingualAE}, the state-of-the-art zero-shot method. In contrast to \qb, we do not observe performance gains as a result of combining different heuristics as UIScore fails to outperform IScore. \ethemes weighted by IScore achieves by far the strongest performance on the hard mentions, despite a relatively poor performance on easy mentions. Overall, the performance makes an encouraging case for the heuristics to be used as strong baselines for entity linking in general, and on large data sets in particular. 

\xhdr{AIDA-CoNLL entity type analysis} The results of the analysis with respect to the entity types available in the original CoNLL 2003 dataset \cite{DBLP:conf/conll/SangM03} are shown in Table~\ref{tab:entity-type}. In CoNLL 2003, there are four entity types: person (PER), organization (ORG), location (LOC), and miscellaneous (MISC).

The UIScore heuristic achieves the best performance on PER mentions, outperforming even m\genre. As described in Subsection \ref{ssec:ecsim}, persons that are mentioned in the news are usually introduced by a simple description of their background or current occupation even if they are well known. Since the heuristics proposed for person disambiguation in \qb are based on this assumption, this explains a relatively strong performance of the UIScore heuristic on PER type entities in AIDA-CoNLL.

Despite superior performance on PER mentions, UIScore lags behind m\genre and \ethemes on other types. We attribute this to a lack of introductory context in comparison to PER mentions (e.g., a mention of ``China'' in an article would typically not be followed by ``a state in East Asia''). Furthermore, non-person named entities are frequently used as metonyms (e.g., ``Kremlin'' is a frequent metonym for the Russian government, but it can also refer to the Kremlin building). Depending on the context, a simple heuristic such as IScore may thus struggle to properly link candidates.

\xhdr{Computational performance}
While m\genre achieves the best performance on both \qb and on AIDA-CoNLL, it is a transformer model and takes \emph{substantially} longer to run in comparison to UIScore. Disambiguating a single mention with m\genre takes approximately 533 times longer than with UIScore, and approximately 533K times longer than with NS, thereby rendering it infeasible for speaker disambiguation in \qb, which contains millions of news articles. For a detailed breakdown of inference times per mention, see Appendix~\ref{app:inference}.


\section{Discussion}

Overall, the results highlight the practicality of the proposed heuristics. Our simple heuristics outperform those based on word embeddings and are competitive in comparison to m\genre.

\subsection{Error analysis}
\label{app:err}
To take a closer at avenues for improvement, we show a manual error analysis for UIScore in Table~\ref{tab:err}. In 6 cases, the predicted entity and the gold entity have a matching domain (e.g., both are sportsmen). In 4 cases, the key property by which a human could determine the correct entity was only implicitly mentioned in the context, which caused a failure in string matching. For 3 articles, a key property of the gold entity was not listed in Wikidata, even though it could be found in external sources such as Wikipedia. The remaining error stems from the presence of a ``decoy'' entity, i.e, an influential but unrelated entity that induced spurious matches. For a thorough description and illustration of the error categories, see Appendix~\ref{app:errdesc}.

\begin{table}[t]
\caption{Error sources for UIScore.}
\label{tab:err}
\centering
\begin{tabular}{lc}
\toprule
             \textbf{Error source}    & \textbf{\#Mentions} \\
\midrule
Similar domain                & 6  (42.9\%)      \\
Key property implicit in the text & 4   (28.6\%)       \\
Key property not in Wikidata  & 3  (21.4\%)       \\
Decoy mention                 & 1  (7.1\%)       \\
\bottomrule
\end{tabular}
\moveup
\moveup
\end{table}

\subsection{Limitations}
Since UIScore is the most promising of our heuristics, we focus on it and its components. 

The biggest limitation of IScore is imposed by the equal importance that is assigned to words in the context, which could be improved by re-ranking important words for given entities. Similarly, Wikidata properties for EEIScore and CSSVE could be ranked or filtered (for example, the property \emph{date of birth} is likely to cause spurious matches, while \emph{occupation} is likely useful). 

Regarding tie-breaking, the use of LQID is intuitive for persons in the news domain, but may fail for other entity types and other domains, and is dependent on Wikidata. Finally, in our focus on \qb data, we are reliant on the authors' method for candidate generation, which could be improved for better performance in the future.


\section{Conclusions and Future Work}

We tackled the problem of entity linking in \qb by employing heuristics that rely on simple signals in the context of mentions and the referent KB. The solid overall performance of the proposed heuristics on \qb, their low computational complexity, and competitive performance on the AIDA-CoNLL benchmark suggest that they can be used as strong baselines for unsupervised entity linking in large datasets.

\xhdr{Future work} 
We plan to experiment with weighting schemes that account for word importance, utilize additional signals from the KB, and include improved candidate generation methods. Finally, we aim to provide a disambiguated version of \qb to the community.

\section*{Acknowledgements}
We would like to thank Vincent Ng for providing insightful feedback during the pre-submission mentorship phase. This project was partly funded by the Swiss National Science Foundation (grant 200021\_185043), the European Union (TAILOR, grant 952215), the Microsoft Swiss Joint Research Center, and the University of Konstanz Zukunfts\-kolleg. We also acknowledge generous gifts from Facebook and Google supporting West’s lab.

\newpage
\balance
\bibliography{nelight}
\bibliographystyle{acl_natbib}

\newpage
\section*{Appendix}
\appendix

\section{Ground Truth Data}
\label{app:gt}
For the method evaluation, we randomly sample 300 articles from \qb. The ground truth for 160 articles is determined by the author, while the remaining 140 articles are annotated by the author's colleagues. The annotators were provided with article content, article title, publication date, article URL, a list of ambiguous named entity mentions, and for each ambiguous mention, a candidate set of QIDs as listed in \qb. The annotators had to either select the correct QID from the candidate set or select one of the following categories if the correct QID is not listed:
\begin{itemize}[leftmargin=*, noitemsep]
   \item \textit{The mention does not refer to a person.} Sometimes, buildings and other artifacts named after some person are identified as a person. We ignore such mentions in the evaluation.
    \item \textit{The correct QID does not exist in Wikidata.} This means that a person is likely not significant enough to have a Wikidata item. For example, sometimes a journalist or a photographer of a newspaper where the article is published shares the name of a famous person and is therefore listed as a speaker candidate. 
    \item \textit{The correct QID exists in Wikidata but is not listed.} This can happen if the correct QID is added to Wikidata after the candidate entities were generated.
    \item \textit{Impossible to determine.} Some articles are either too noisy or do not contain enough information for disambiguation to be feasible.
\end{itemize}

In Table \ref{tab:gt}, we present the distribution of person mentions in the evaluation data with respect to different categories. We observe that more than 70\% of the 1866 mentions are unambiguous. For 310 (57\%) of the ambiguous mentions, it was possible to determine the ground truth based on the given candidate sets. For the majority of the remaining 43\% of ambiguous mentions no correct entity was available in Wikidata.

The main drawback of the \qb evaluation dataset is its small size. Since all articles were annotated by only one annotator, there is no data on the inter-annotator agreement. In the future, we aim to create a more sophisticated benchmark dataset via crowdsourcing.

\begin{table}[!ht]
\caption{Distribution of mentions in the ground truth data with respect to ambiguity and availability of ground truth.}
\label{tab:gt}
\resizebox{\linewidth}{!}{
\begin{tabular}{llc}
\toprule
\multicolumn{2}{l}{\textbf{Category}}                            & \textbf{\#Mentions}   \\ \midrule
\multicolumn{2}{l}{Unambiguous}                         & 1322 (70.8\%) \\ \midrule
\multirow{5}{*}{Ambiguous} & Gold entity exists         & 310 (16.6\%)  \\
                           & No correct QID in Wikidata & 151 (8.1\%)   \\
                           & Impossible                 & 37 (2.0\%)    \\
                           & Correct QID not listed     & 24 (1.3\%)    \\
                           & Not a person               & 22 (1.2\%)    \\ \midrule
\multicolumn{2}{l}{Total}                               & 1866          \\ \bottomrule
\end{tabular}}
\moveup
\moveup
\moveups
\end{table}

\section{Implementation Details of the Scoring Methods}
 \label{app:methods}
\subsection{IScore}
To calculate the IScore, we first obtain labels of Wikidata statement values listed $e$. We then tokenize the content of $a$ using the tagset of the Penn Treebank Tokenizer. We use the computed tokens to create sets $\mathcal W_a$ and $\mathcal W_e$. Then, we apply the formula given in equation \ref{eq:iscore} and compute the IScore based on $\mathcal W_a$, $\mathcal W_e$, and a predefined set of English stopwords $\mathcal W_{sw}$\footnote{\href{https://gist.github.com/sebleier/554280}{https://gist.github.com/sebleier/554280}}.

\subsection{CSE}
\label{app:cse}
To embed an article, we follow the standard transformer model preprocessing procedure. We tokenize the article content using the model-specific tokenizer, respecting BART's 1024 token limit by simply truncating the input if the limit is exceeded. We then feed the obtained tokens to BART and average the last hidden state of the model output. Since truncation leads to loss of information in comparison to other methods, we experimented with chunking the input into chunks of at most 1024 tokens, computing token embeddings in each chunk separately, and aggregating the obtained token embeddings. However, this did not improve performance on \qb (0.698 P@1 and 0.818 MRR), while all articles from AIDA-CoNLL are within the token limit so we report the results of the first approach.

Embedding the entity is slightly more challenging. Following the same procedure as for the computation of the article content embeddings, we compute the embedding the the first paragraph in an entity's Wikipedia page if such a page is available. Otherwise, we compute the embeddings of the short description, and each statement value label listed for an entity in Wikidata, and aggregate them via arithmetic mean.

\subsection{mGENRE}
\label{app:mgenre}
We use m\genre in a similar setup as \citet{DeCao2021MultilingualAE}. Suppose that we want to disambiguate entity mention $m$ occurring in an article $a$. We first enclose $m$ with special tokens \texttt{[START]} and \texttt{[END]} that correspond to the start and the end of a mention span. We then take at most $t$ mBART \cite{liu-etal-2020-multilingual-denoising} tokens from either side. As the input for m\genre, we use a string consisting of the left context, the mention enclosed with the special tokens, and the right context. m\genre then outputs the top $k$ entity QIDs and their respective scores, where $k$ is the beam size. For entities in $\mathcal Q_m$ that are not retrieved by m\genre, we simply assign 0 as a score. Note that m\genre outputs the scores corresponding to the negative log-likelihood of the resulting sequence. Thus, in order for 0 to be the smallest possible score, we exponentiate the scores obtained from m\genre. In the \qb setup, we also perform one additional step: since each speaker candidate can be mentioned multiple times in the text, we run m\genre for each of the speaker candidate mentions and sum the scores obtained for each of the candidate Wikidata entities.

In Table \ref{tab:mgenre-context}, we present the performances of m\genre on both \qb and AIDA-CoNLL for different values of $t$, while in Table \ref{tab:main-res} we report only the best obtained P@1. In all our experiments with m\genre, we set the beam size $k$ to 10.

\begin{table}[t]
\caption{Performance of m\genre for different context sizes. The best result in each column is highlighted \textbf{bold}.}
\centering
\resizebox{.99\linewidth}{!}{
\begin{tabular}{lcccc}
\hline
    & \multicolumn{2}{c}{\textbf{\qb}} & \multicolumn{2}{c}{\textbf{AIDA-CoNLL}} \\ \cmidrule(lr){2-3}\cmidrule(lr){4-5} 
$t$ & \textbf{P@1}       & \textbf{MRR}      & \textbf{P@1}       & \textbf{MRR}       \\ \hline
64  & 0.951 $\pm$ 0.029  & 0.968 $\pm$ 0.018 & 0.664 $\pm$ 0.013  & 0.713 $\pm$ 0.012  \\
128 & \textbf{0.963 $\pm$ 0.025}  & \textbf{0.976 $\pm$ 0.017} & 0.675 $\pm$ 0.014  & 0.723 $\pm$ 0.013  \\
256 & 0.959 $\pm$ 0.026  & 0.972 $\pm$ 0.021 & \textbf{0.682 $\pm$ 0.014}  & \textbf{0.730 $\pm$ 0.013}  \\ \hline
\end{tabular}}
\moveup
\moveup
\moveups
\label{tab:mgenre-context}
\end{table}

\section{Evaluation Setup Details} 
\label{app:qb-aida-diffs}
\xhdr{\qb} The \qb data exclusively contains annotations of person mentions. Before training a model that attributes the quotations to their respective speakers, the quotations and speaker candidates are identified in the article text \cite{quotebank}. The extraction of speaker candidates is explained in detail by \citet{quootstrap}. Although a speaker candidate can appear in an article multiple times, the quotations are not attributed to specific mentions but rather to the most likely speaker candidate. Thus, we evaluate our methods on \qb on a speaker candidate level and refer to speaker candidates as mentions to ensure that our method and result descriptions are consistent with the standard nomenclature.

\xhdr{AIDA-CoNLL} When evaluating our methods on the AIDA-CoNLL benchmark, we do not ignore the mentions for which the gold entity either cannot be determined or is not retrieved by the candidate generator. As a consequence, the resulting P@1 and MRR reported on AIDA-CoNLL are significantly lower in comparison to the \qb results as they are bounded by the recall of the candidate generator. We use the same candidate generator as \citet{arora-etal-2021-low}, which imposes an upper bound of 0.824 to P@1 and MRR. Additionally, to ensure a fair comparison with \citet{arora-etal-2021-low}, we break ties by selecting the first speaker candidate with the same score and use the same definition of the easy and hard mentions when reporting the method performances.

\section{Wikidata}
\label{app:wiki}
Wikidata is a large community-driven KB. It boasts more than 96 million data items as of January 2022, out of which ~6 million are humans. Each Wikidata item is identified by a unique positive integer prefixed with the upper-case letter Q, also known as QID (e.g. Earth (Q2), Mahatma Gandhi (Q1001)). Obligatory data fields of items are a label and a description. Labels and descriptions need not be unique, but each item is uniquely identified by a combination of a label and a short description. Therefore, each QID is linked to the label-description combination. Optionally, some items consist of aliases (alternative names for an entity) and statements. Statements provide additional information about an item and they consist of at least one property-value pair. A property is a pre-defined data type, identified by a unique positive integer, but unlike items, it is prefixed with the upper-case letter P (e.g. occupation (P106), sex or gender (P21)). The value of a statement may take on many types, such as Wikidata items, strings, numbers, or media files. Some items also have a list of site links that connect them to the corresponding page of the entity in other Wikimedia projects, such as Wikipedia or Wikibooks. The methods we propose in Section \ref{sec:scoring} leverage the described information to link the named entity mentions in the news articles to their respective Wikidata entities.

\begin{table*}[ht]
\caption{Results of the IScore ablation study with respect to word normalization and inclusion of different Wikidata features. In each row, we report P@1 and MRR of IScore method for the combinations of the following Wikidata features: short description (D), Wikipedia first paragraph (P), statement value labels (S), and statement value labels and aliases ($\text{S}_\text{A}$) for a setting without word normalization, as well as for settings with stemming and lemmatization. The best results in each column are in bold. Since $\text{S}_\text{A}$ is essentially a superset of S, we omit the combinations where both S and $\text{S}_\text{A}$ appear. All the experiments were run with NS as a tie-breaker.}
\label{tab:iscoreabl}
\resizebox{\textwidth}{!}{
\begin{tabular}{lcccccc}
\toprule
                     & \multicolumn{2}{c}{\textbf{No normalization}}       & \multicolumn{2}{c}{\textbf{Lemmatization}}          & \multicolumn{2}{c}{\textbf{Stemming}}               \\ \cmidrule(lr){2-3}\cmidrule(lr){4-5}\cmidrule(lr){6-7}
\textbf{Combination} & \textbf{P@1}             & \textbf{MRR}             & \textbf{P@1}             & \textbf{MRR}             & \textbf{P@1}             & \textbf{MRR}             \\ \midrule
D           & 0.869 $\pm$ 0.044          & 0.921 $\pm$ 0.027          & 0.890 $\pm$ 0.040          & 0.930 $\pm$ 0.026          & 0.894 $\pm$ 0.039          & 0.934 $\pm$ 0.026          \\
P           & 0.832 $\pm$ 0.049          & 0.903 $\pm$ 0.030          & 0.816 $\pm$ 0.051          & 0.895 $\pm$ 0.029         & 0.832 $\pm$ 0.047          & 0.902 $\pm$ 0.029          \\
S           & 0.894 $\pm$ 0.040          & 0.936 $\pm$ 0.026          & 0.898 $\pm$ 0.039          & 0.940 $\pm$ 0.024          & 0.906 $\pm$ 0.038          & 0.944 $\pm$ 0.024          \\
$\text{S}_\text{A}$           & 0.886 $\pm$ 0.041          & 0.932 $\pm$ 0.025          & 0.890 $\pm$ 0.042          & 0.935 $\pm$ 0.025          & 0.898 $\pm$ 0.039          & 0.939 $\pm$ 0.024          \\ \midrule
D + P       & 0.841 $\pm$ 0.046          & 0.907 $\pm$ 0.028          & 0.820 $\pm$ 0.050          & 0.898 $\pm$ 0.030         & 0.841 $\pm$ 0.047          & 0.906 $\pm$ 0.028          \\
D + S       & \textbf{0.902 $\pm$ 0.039} & \textbf{0.943 $\pm$ 0.024} & \textbf{0.906 $\pm$ 0.038} & 0.945 $\pm$ 0.022 & \textbf{0.918 $\pm$ 0.035} & \textbf{0.952 $\pm$ 0.021} \\
D + $\text{S}_\text{A}$ & 0.890 $\pm$ 0.041 & 0.937 $\pm$ 0.023 & \textbf{0.906 $\pm$ 0.038} & \textbf{0.947 $\pm$ 0.022} & 0.914 $\pm$ 0.037 & 0.950 $\pm$ 0.022 \\
P + S       & 0.861 $\pm$ 0.044          & 0.919 $\pm$ 0.028          & 0.861 $\pm$ 0.045          & 0.920 $\pm$ 0.028          & 0.873 $\pm$ 0.044          & 0.925 $\pm$ 0.026          \\
 P + $\text{S}_\text{A}$       & 0.878 $\pm$ 0.042          & 0.928 $\pm$ 0.025          & 0.882 $\pm$ 0.041          & 0.931 $\pm$ 0.025          & 0.882 $\pm$ 0.042          & 0.930 $\pm$ 0.025          \\ \midrule

D + P + S   & 0.861 $\pm$ 0.045 & 0.921 $\pm$ 0.026 & 0.861 $\pm$ 0.045 & 0.920 $\pm$ 0.027 & 0.873 $\pm$ 0.042 & 0.926 $\pm$ 0.026 \\
 D + P + $\text{S}_\text{A}$   & 0.886 $\pm$ 0.042 & 0.934 $\pm$ 0.025 & 0.886 $\pm$ 0.041 & 0.934 $\pm$ 0.025 & 0.882 $\pm$ 0.042 & 0.930 $\pm$ 0.026 \\ \bottomrule
\end{tabular}}
\moveup
\end{table*}

\begin{table}[ht]
\caption{Comparison of performances of CSE and IScore when considering different context sizes. Ensemble refers to the sum of the scores obtained considering the narrow and entire context of the article, respectively. The best results for each scoring method are in bold. All the experiments were run with NS as a tie-breaker.}
\centering
\label{tab:con}
\resizebox{.48\textwidth}{!}{
\begin{tabular}{llcc}
\toprule
\textbf{Method}                  & \textbf{Context}  & \textbf{P@1}    & \textbf{MRR}    \\ \midrule
\multirow{3}{*}{CSE}    & Narrow   & 0.751 $\pm$ 0.055 & 0.857 $\pm$ 0.033 \\
                                 & Entire   & 0.833 $\pm$ 0.050 & 0.902 $\pm$ 0.029 \\
                                 & Ensemble & \textbf{0.857 $\pm$ 0.044} & \textbf{0.921 $\pm$ 0.025} \\ \midrule
\multirow{3}{*}{IScore} & Narrow   & 0.898 $\pm$ 0.039 & 0.941 $\pm$ 0.023 \\
                                 & Entire   & 0.918 $\pm$ 0.035 & 0.952 $\pm$ 0.021 \\
                                 & Ensemble & \textbf{0.922 $\pm$ 0.036} & \textbf{0.954 $\pm$ 0.022} \\ \bottomrule
\end{tabular}}
\moveup
\end{table}

\section{Additional Experiments}
\label{app:iscoreexp}

\subsection{Wikidata Features and Word Normalization Ablation for IScore}
\label{app:wikinorm}
In Table \ref{tab:iscoreabl}, we show the results of an ablation study that aims to assess the effect of the inclusion of different Wikidata entity features on the performance of IScore and word normalization methods. The features we consider are short descriptions, statement value labels with and without aliases, and Wikipedia first paragraphs. We obtain the best results by leveraging short descriptions and Wikidata statement values. When using only Wikipedia first paragraphs, we obtain a performance similar to NS, a simple entity popularity metric. Seemingly, the inclusion of aliases does not improve the performance. Additionally, we observe that lemmatization (using the WordNet lemmatizer \cite{wordnet}) and stemming (using the Porter stemmer \cite{porter1980algorithm}) improve IScore performance by a small margin. Furthermore, we observe a slight performance gain of stemming over lemmatization. This is especially important considering the volume of the data and the inefficiency of lemmatization when compared to stemming.

\subsection{Context Size}
As shown in Table \ref{tab:con}, narrowing down the context has a negative impact on the performances of both the CSE and IScore scoring methods. However, we hypothesize that the words that occur close to the entity mention are more important than those in a broader context. Therefore, we also experiment with the linear combination of the respective scores for each context size. In both cases, the optimal weights obtained through grid search optimization are $(1, 1)$. We observe a slight performance gain for the ensemble of both scoring methods.

\subsection{Tie breakers}
\label{app:tie}
In Table~\ref{tab:tie}, we present the results of the experiment with various tiebreakers. Seemingly, all the tie-breakers are a reasonable choice since no tie-breaker clearly outperforms the others.

\begin{table*}[ht]
\caption{P@1 of different popularity metrics as tiebreakers. Rows correspond to scoring methods and columns to tiebreakers. CSE and UCSE are omitted from the table because their performance remains the same irrespective of the tiebreaker. The best P@1 in each row is highlighted \textbf{bold}.}
\label{tab:tie}
\resizebox{.99\linewidth}{!}{
\begin{tabular}{lcccccc}
\toprule
                  & \textbf{NS}    & \textbf{NP}    & \textbf{PR WP} & \textbf{PR WD} & \textbf{LQID}  \\ \hline
\textbf{IScore}   & 0.918 $\pm$ 0.036       & \textbf{0.922 $\pm$ 0.035} & 0.918 $\pm$ 0.036        & 0.918  $\pm$ 0.036        & 0.906  $\pm$ 0.038        \\
\textbf{EEIScore} & 0.898 $\pm$ 0.039         & 0.894 $\pm$ 0.039 & \textbf{0.906 $\pm$ 0.037}          & 0.878  $\pm$ 0.042         & 0.873 $\pm$ 0.042          \\
\textbf{CSSVE}    & \textbf{0.784 $\pm$ 0.052} & 0.780 $\pm$ 0.054        & \textbf{0.784 $\pm$ 0.053} & \textbf{0.784 $\pm$ 0.051} &  \textbf{0.784 $\pm$ 0.052} \\
\textbf{UIScore}  & 0.939  $\pm$ 0.032        & 0.939 $\pm$ 0.032        & \textbf{0.942 $\pm$ 0.031} & 0.935  $\pm$ 0.033        & 0.931  $\pm$ 0.033  \\
\bottomrule
\end{tabular}}
\moveup
\moveups
\end{table*}

\section{Inference Time}
\label{app:inference}
\begin{table}[ht]
\caption{Estimated per-mention inference times of the selected methods. m\genre was executed on Nvidia GeForce GTX TITAN X, while UIScore and NS were executed on a single 2.5 GHz core of Intel Xeon E5-2680 processor.}
\label{tab:eff}
\centering
\begin{tabular}{lccc}
\toprule
        & \multicolumn{2}{c}{Inference time} \\ \cline{2-3} 
Method  & \qb             & AIDA-CoNLL       \\ \midrule
m\genre  & 8.0\ s         & 1.9\ s             \\
NS      & 15 $\mu$s       & 26 $\mu$s         \\
IScore  & 7.9 ms          & 67 ms            \\
UIScore & 15 ms           & 135 ms           \\
Eigen   & 11 ms           & 39 ms \\ \bottomrule
\end{tabular}
\moveup
\end{table}

In Table \ref{tab:eff}, we present the inference times of m\genre, \ethemes, our best-performing methods on \qb and AIDA-CoNLL: UIScore and IScore, respectively, and the well-performing entity popularity metric NS. \ethemes and the selected heuristics are significantly more efficient than m\genre. The differences in inference times on Quotebank and AIDA-CoNLL are due to the setup differences (see \ref{app:qb-aida-diffs}). Additionally, the inference times of NS, IScore, UIScore, and \ethemes largely depend on the number of candidates per mention. Thus, since on average, the number of candidate entities per mention on AIDA-CoNLL (approx. 18) is substantially larger than in \qb (approx. 5), their inference times on AIDA-CoNLL are longer. Note that our best methods do not require GPU, making them easily parallelizable on CPU cores. 

\section{Mean reciprocal rank of the methods}
\label{app:main-mrr}
As an extension of Table \ref{tab:main-res},  in Table \ref{tab:main-mrr} we present the MRR of the methods. MRR follows similar trends as P@1. 

\begin{table*}[ht]
\tabcaption{MRR of the methods on \qb and AIDA-CoNLL. Eigen and Eigen (IScore) have the same definition as in Table \ref{tab:main-res}. The best obtained MRR in each column is highlighted \textbf{bold}.}
\label{tab:main-mrr}
\vspace*{2mm}
\centering
\begin{threeparttable}
\resizebox{\textwidth}{!}{
\begin{tabular}{lcccccc}
\hline
                   & \multicolumn{3}{c}{\textbf{\qb}}                                               & \multicolumn{3}{c}{\textbf{AIDA-CoNLL}}                                              \\ \cmidrule(lr){2-4}\cmidrule(lr){5-7} 
                   & \textbf{Easy}              & \textbf{Hard}              & \textbf{Overall}           & \textbf{Easy}              & \textbf{Hard}              & \textbf{Overall}           \\ \hline
Random             & 0.622 $\pm$ 0.022          & 0.484 $\pm$ 0.058          & 0.597 $\pm$ 0.030          & 0.387 $\pm$ 0.013          & 0.205 $\pm$ 0.006          & 0.273 $\pm$ 0.009          \\
 \midrule
LQID               & 0.904 $\pm$ 0.030          & 0.505 $\pm$ 0.094          & 0.836 $\pm$ 0.036          & 0.912 $\pm$ 0.009          & 0.451 $\pm$ 0.021          & 0.635 $\pm$ 0.013          \\
NP                 & 0.959 $\pm$ 0.021          & 0.457 $\pm$ 0.082          & 0.873 $\pm$ 0.034          & 0.901 $\pm$ 0.010          & 0.352 $\pm$ 0.021          & 0.603 $\pm$ 0.013          \\
NS                 & \textbf{1.000 $\pm$ 0.000} & 0.389 $\pm$ 0.044          & 0.895 $\pm$ 0.031          & 0.943 $\pm$ 0.007          & 0.485 $\pm$ 0.020          & 0.661 $\pm$ 0.012          \\
PR$_{\mathrm{WD}}$ & 0.873 $\pm$ 0.032          & 0.453 $\pm$ 0.098          & 0.801 $\pm$ 0.039          & 0.903 $\pm$ 0.009          & 0.336 $\pm$ 0.019          & 0.601 $\pm$ 0.013          \\
PR$_{\mathrm{WP}}$ & 0.962 $\pm$ 0.020          & 0.561 $\pm$ 0.101          & 0.893 $\pm$ 0.031          & \textbf{0.966 $\pm$ 0.005} & 0.491 $\pm$ 0.020          & 0.676 $\pm$ 0.012          \\ \midrule
IScore             & 0.977 $\pm$ 0.016          & 0.842 $\pm$ 0.096          & 0.954 $\pm$ 0.022          & 0.908 $\pm$ 0.009          & 0.686 $\pm$ 0.021          & 0.692 $\pm$ 0.013          \\
NIScore            & 0.980 $\pm$ 0.016          & 0.750 $\pm$ 0.093          & 0.941 $\pm$ 0.023          & 0.903 $\pm$ 0.010          & 0.538 $\pm$ 0.024          & 0.651 $\pm$ 0.013          \\
CSE                & 0.947 $\pm$ 0.023          & 0.682 $\pm$ 0.099          & 0.902 $\pm$ 0.029          & 0.871 $\pm$ 0.011          & 0.455 $\pm$ 0.024          & 0.612 $\pm$ 0.013          \\ \midrule
EEIScore           & 0.972 $\pm$ 0.018          & 0.801 $\pm$ 0.097          & 0.943 $\pm$ 0.023          & 0.555 $\pm$ 0.016          & 0.467 $\pm$ 0.022          & 0.435 $\pm$ 0.011          \\
CSSVE              & 0.930 $\pm$ 0.027          & 0.586 $\pm$ 0.100          & 0.871 $\pm$ 0.033          & 0.796 $\pm$ 0.013          & 0.412 $\pm$ 0.023          & 0.559 $\pm$ 0.013          \\ \midrule
UIScore            & 0.980 $\pm$ 0.015          & \textbf{0.891 $\pm$ 0.080} & 0.965 $\pm$ 0.019          & 0.888 $\pm$ 0.010          & 0.718 $\pm$ 0.020          & 0.689 $\pm$ 0.013          \\
UCSE               & 0.970 $\pm$ 0.018          & 0.743 $\pm$ 0.099          & 0.931 $\pm$ 0.025          & 0.874 $\pm$ 0.011          & 0.630 $\pm$ 0.021          & 0.659 $\pm$ 0.013          \\ \midrule
Eigen (IScore)     & 0.974 $\pm$ 0.018          & 0.817 $\pm$ 0.092          & 0.947 $\pm$ 0.024          & 0.864 $\pm$ 0.011          & \textbf{0.804 $\pm$ 0.020}\tnote{\dag} & 0.697 $\pm$ 0.013          \\
Eigen              & 0.998 $\pm$ 0.005          & 0.529 $\pm$ 0.090          & 0.917 $\pm$ 0.027          & 0.910 $\pm$ 0.009          & 0.674 $\pm$ 0.019          & 0.690 $\pm$ 0.012          \\
m\genre             & 0.998 $\pm$ 0.005          & 0.869 $\pm$ 0.089          & \textbf{0.976 $\pm$ 0.017} & 0.959 $\pm$ 0.006          & 0.720 $\pm$ 0.022          & \textbf{0.730 $\pm$ 0.012}\tnote{\dag} \\ \bottomrule
\end{tabular}}
\begin{tablenotes}
\small \item[\dag] Indicates statistical significance ($p<0.05$) between the best and the second-best method using bootstrapped 95\% CIs.
\end{tablenotes}
\end{threeparttable}
\moveup
\moveups
\end{table*}

\section{Error Source Descriptions}
\label{app:errdesc}

\xhdr{Similar domain} If the gold entity and the system output have similar backgrounds or occupations, their Wikidata items tend to contain similar statements. For example, in one of the articles, the gold entity for Shawn Williams was Q7491485 (lacrosse player), while the output of the model was Q13064143 (American football player, defensive back). Shawn Williams first appears in the following sentence:\\[3pt]
\textit{Canada head coach Randy Mearns kept his No. 51 warm-up shirt - honoring Tucker Williams, the son of NLL star \textbf{\underline{Shawn Williams}} of the Buffalo Bandits who is currently undergoing the treatment for Burkitt's Lymphoma - on throughout the game.}\\[3pt]
Earlier in the article, \textit{lacrosse} was mentioned directly, which in addition to the mention of \textit{NLL} (National Lacrosse League) made it clear that Q7491485 is the gold entity. However, the UIScore of Q13064143 was just 1 point higher than the UIScore of Q7491485, which led to the erroneous prediction.

\xhdr{Key property not in Wikidata} In some cases, the Wikidata item does not contain the key information that is used to describe the entity in the article. Such cases are difficult even for humans as they require background knowledge stored in multiple sources. An example of this is John Prendergast (Q6253345), who was described in one article as the \textit{co-founder of Enough}. This property is not listed in the Wikidata item of Q6253345 but can be found in external sources. The output of the model was Q6253343, a late British Army officer who served in World War II. The article in which Prendergast was mentioned was about violent events in Congo and was thus rich in war-related terms. Most importantly, World War II was mentioned in the article, leading to three spuriously matched words in Q6253343's Wikidata item. The final scores of Q6253345 and Q6253343 were 8 and 12 respectively. If \textit{co-founder of Enough} was listed in Wikidata and if \textit{World War II} was treated as a single noun phrase, the UIScore of the gold entity, Q6253345, would beat the score of Q6253343.

\xhdr{Key property implicit in text} Some errors occur when enough information is provided in the article and in Wikidata, but the key properties are not mentioned in the text explicitly. For example, professional golfer Will Mackenzie (Q8002946) was mentioned in an article that was clearly about golf. However, golf was not mentioned at all in the article, yet Mackenzie's profession could be inferred from other terms related to golf, such as PGA Tour, which does not appear in the Wikidata item of Q8002946. The output of the method was Q4019878 (actor and director). Although there were other golfers mentioned in the article (leading to an EEIScore of 4 for Q8002946), its item matched no stems in text, while Q4019878 matched two stems that were completely unrelated to the article: \textit{provid} (He was born in Providence which shares the same stem as provide) and \textit{televis} (he was a television actor). Furthermore, Q4019878 matched citizenship, spoken language, and gender with other unambiguous mentions in the article. As a result, Q4019878 was the predicted label. This indicates the need for assigning weights to Wikidata properties to avoid irrelevant matches.

\xhdr{Decoy mention} To illustrate the decoy mention error source, we consider the following example:\\[3pt]
\textit{"Amazon will debut five new comedy drama pilots in 2014, including "The After", from \textbf{\underline{Chris Carter}} ("The X-Files"); "Bosch", based on book series by Michael Conelly; "Mozart in the Jungle", from Roman Coppola ("The Darjeeling Limited"); "The Rebels" from former New York Giants football player \textbf{\underline{Michael Strahan}}; and "Transparent" from Jill Soloway ("Six Feet Under")."}\\[3pt]
Suppose that we want to disambiguate Chris Carter. Clearly, the correct entity corresponding to Chris Carter is the movie producer who created the science-fiction drama "The X-Files" (Q437267). However, the appearance of Michael Strahan increased the IScore of sportsmen named Chris Carter that played for a New York team (due to the appearance of the words "player", "New", and "York"). Note that a limitation of IScore is that it treats the words New and York separately, although they should be treated as a single noun phrase.

\end{document}

%% file: dlab_macros.tex
\usepackage[utf8]{inputenc}
\usepackage[T1]{fontenc}
\usepackage{hyphenat}
\usepackage{xspace}
\usepackage{amsmath}
\usepackage{amsfonts}
\usepackage{hyperref}
\usepackage{url}
\usepackage{booktabs}
\usepackage{multirow}
\usepackage{makecell}
\usepackage{caption}
\usepackage{minibox}
\usepackage{bbm}
\usepackage{graphicx}
\usepackage{balance}
\usepackage{mathtools}
\usepackage{color}
\usepackage{marvosym}
\usepackage{ifthen}
\usepackage{textcomp}
\usepackage{enumitem}
\usepackage{verbatim}
\usepackage{algorithm}
\usepackage{numprint}
\usepackage{balance}

\usepackage{amsthm}
\theoremstyle{plain}

\newcommand{\chatoDisplayMode}[1]{#1}

\definecolor{MyRed}{rgb}{0.6,0.0,0.0} 
\definecolor{MyBlack}{rgb}{0.1,0.1,0.1} 
\newcommand{\inred}[1]{{\color{MyRed}\sf\textbf{\textsc{#1}}}}
\newcommand{\frameit}[2]{
  \begin{center}
  {\color{MyRed}
  \framebox[.9\columnwidth][l]{
    \begin{minipage}{.85\columnwidth}
    \inred{#1}: {\sf\color{MyBlack}#2}
    \end{minipage}
  }\\
  }
  \end{center}
}

\newcommand{\note}[2][]{\chatoDisplayMode{\def\@tmpsig{#1}\frameit{{\Pointinghand} Note}{#2\ifx \@tmpsig \@empty \else \mbox{ --\em #1}\fi}}}
\newcommand{\todo}[2][]{\chatoDisplayMode{\def\@tmpsig{#1}\frameit{{\Writinghand} To-do}{#2\ifx \@tmpsig \@empty \else \mbox{ --\em #1}\fi}}}

\newcommand{\abbrevStyle}[1]{#1}

\newcommand{\ie}{\abbrevStyle{i.e.}\xspace}

\newcommand{\etc}{\abbrevStyle{etc.}\xspace}

\newcommand{\xhdr}[1]{\vspace{1.7mm}\noindent{{\bf #1.}}}

\newcommand{\xhdrNoPeriod}[1]{\vspace{1.7mm}\noindent{{\bf #1}}}

\newcommand{\textcite}[1]{\citeauthor{#1} \shortcite{#1}}

\newcommand{\hide}[1]{}

\makeatletter
\newcommand{\iffont}[2]{\ifthenelse{\equal{\f@family}{#1}}{#2}{}}
\makeatother

  \usepackage{mathptmx}

  \DeclareSymbolFont{greek}{OML}{cmm}{m}{n}
  \DeclareMathSymbol{\alpha}{\mathalpha}{greek}{"0B}
  \DeclareMathSymbol{\beta}{\mathalpha}{greek}{"0C}
  \DeclareMathSymbol{\gamma}{\mathalpha}{greek}{"0D}
  \DeclareMathSymbol{\delta}{\mathalpha}{greek}{"0E}
  \DeclareMathSymbol{\epsilon}{\mathalpha}{greek}{"0F}
  \DeclareMathSymbol{\zeta}{\mathalpha}{greek}{"10}
  \DeclareMathSymbol{\eta}{\mathalpha}{greek}{"11}
  \DeclareMathSymbol{\theta}{\mathalpha}{greek}{"12}
  \DeclareMathSymbol{\iota}{\mathalpha}{greek}{"13}
  \DeclareMathSymbol{\kappa}{\mathalpha}{greek}{"14}
  \DeclareMathSymbol{\lambda}{\mathalpha}{greek}{"15}
  \DeclareMathSymbol{\mu}{\mathalpha}{greek}{"16}
  \DeclareMathSymbol{\nu}{\mathalpha}{greek}{"17}
  \DeclareMathSymbol{\xi}{\mathalpha}{greek}{"18}
  \DeclareMathSymbol{\pi}{\mathalpha}{greek}{"19}
  \DeclareMathSymbol{\rho}{\mathalpha}{greek}{"1A}
  \DeclareMathSymbol{\sigma}{\mathalpha}{greek}{"1B}
  \DeclareMathSymbol{\tau}{\mathalpha}{greek}{"1C}
  \DeclareMathSymbol{\upsilon}{\mathalpha}{greek}{"1D}
  \DeclareMathSymbol{\phi}{\mathalpha}{greek}{"1E}
  \DeclareMathSymbol{\chi}{\mathalpha}{greek}{"1F}
  \DeclareMathSymbol{\psi}{\mathalpha}{greek}{"20}
  \DeclareMathSymbol{\omega}{\mathalpha}{greek}{"21}
  \DeclareMathSymbol{\varepsilon}{\mathalpha}{greek}{"22}
  \DeclareMathSymbol{\vartheta}{\mathalpha}{greek}{"23}
  \DeclareMathSymbol{\varpi}{\mathalpha}{greek}{"24}
  \DeclareMathSymbol{\varrho}{\mathalpha}{greek}{"25}
  \DeclareMathSymbol{\varsigma}{\mathalpha}{greek}{"26}
  \DeclareMathSymbol{\varphi}{\mathalpha}{greek}{"27}
  \DeclareSymbolFont{otone}{OT1}{cmr}{m}{n}
  \DeclareMathSymbol{\Gamma}{\mathalpha}{otone}{0}
  \DeclareMathSymbol{\Delta}{\mathalpha}{otone}{1}
  \DeclareMathSymbol{\Theta}{\mathalpha}{otone}{2}
  \DeclareMathSymbol{\Lambda}{\mathalpha}{otone}{3}
  \DeclareMathSymbol{\Xi}{\mathalpha}{otone}{4}
  \DeclareMathSymbol{\Pi}{\mathalpha}{otone}{5}
  \DeclareMathSymbol{\Sigma}{\mathalpha}{otone}{6}
  \DeclareMathSymbol{\Upsilon}{\mathalpha}{otone}{7}
  \DeclareMathSymbol{\Phi}{\mathalpha}{otone}{8}
  \DeclareMathSymbol{\Psi}{\mathalpha}{otone}{9}
  \DeclareMathSymbol{\Omega}{\mathalpha}{otone}{10}
  \DeclareSymbolFont{syms}{OML}{cmm}{m}{it}
  \DeclareMathSymbol{\partial}{\mathord}{syms}{"40}
  \DeclareMathAlphabet{\mathbold}{OML}{cmm}{b}{it}
  \DeclareSymbolFont{largesymbols}{OMX}{cmex}{m}{n}
